# EasyCalib: Simple and Low-Cost In-Situ Calibration for Force Reconstruction with Vision-Based Tactile Sensors

Mingxuan Li, *Graduate Student Member, IEEE*, Lunwei Zhang, *Graduate Student Member, IEEE*, Yen Hang Zhou, *Graduate Student Member, IEEE*, Tiemin Li, and Yao Jiang, *Member, IEEE*

*Abstract*—For elastomer-based tactile sensors, represented by visuotactile sensors, routine calibration of mechanical parameters (Young's modulus and Poisson's ratio) has been shown to be important for force reconstruction. However, the reliance on existing in-situ calibration methods for accurate force measurements limits their cost-effective and flexible applications. This article proposes a new in-situ calibration scheme that relies only on comparing contact deformation. Based on the detailed derivations of the normal contact and torsional contact theories, we designed a simple and low-cost calibration device, EasyCalib, and validated its effectiveness through extensive finite element analysis. We also explored the accuracy of EasyCalib in the practical application and demonstrated that accurate contact distributed force reconstruction can be realized based on the mechanical parameters obtained. EasyCalib balances low hardware cost, ease of operation, and low dependence on technical expertise and is expected to provide the necessary accuracy guarantees for wide applications of visuotactile sensors in the wild.

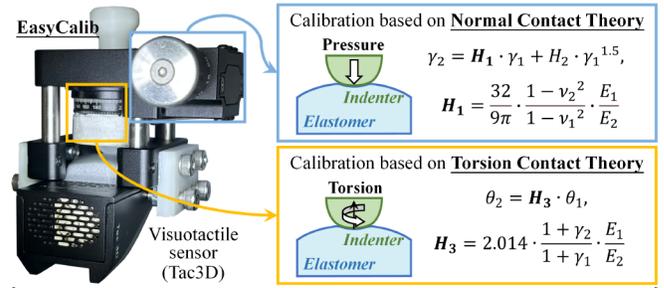

**Fig. 1.** EasyCalib: In-situ mechanical calibration device and method.

## I. Introduction

### A. Background

Tactile perception can help robots obtain the properties, contact status, and motion trends of current interacting objects, which is crucial for in-hand manipulation and tool usage [1], [2]. In recent years, visuotactile sensors [3], [4] (or vision-based tactile sensors) have received widespread attention from the robot community. Such sensors can measure the deformation of the contact elastomer through visual methods, and convert tactile features into high-resolution and multimodal tactile images. Due to its ability to provide accurate information on the contact force and its distribution on the robot's fingertips, such sensors have been applied in various tasks, including grasping force control [5], dexterous operation [6], and swing-up control [7].

For visuotactile sensors, obtaining reliable mechanical parameters of the contact elastomer (including Young's modulus $E$ and Poisson's ratio $\nu$) has been proven to be crucial for measuring contact force distribution [8]. The GelSlim sensor used Young's modulus and Poisson's ratio to obtain the stiffness matrix of the gel pad, and determined the applied force based on the standard Finite Element Method (FEM) [9]. Sferrazza *et al.* constructed the relationship between the contact force and marker displacement based on the Boussinesq-Cerritti equation, which could be described as a matrix mapping containing these two parameters [10]. Zhang *et al.* obtained the inverse mapping model from the measured displacement to the estimated distributed force using FEM analysis software, which relies on the set of $E$ and $\nu$ to derive the total stiffness matrix [11].

### B. Related Work

Due to manufacturing errors, it is difficult to ensure that each sensor's elastomer has the same mechanical parameters. To ensure reliable force reconstruction, it is necessary to effectively measure Young's modulus and Poisson's ratio of each sensor's elastomer. Meanwhile, relevant studies have shown that the mechanical parameters of elastomers undergo significant changes under long-term use [12]. Such issues require the user to perform mechanical calibration on the elastomer regularly.

Zhao *et al.* first proposed an in-situ calibration method suitable for visuotactile sensors [8]. This method applies indentation testing with a cylindrical indenter to obtain force/deformation curves, and then calculates Young's modulus and Poisson's ratio based on the referenced contact model. Compared to traditional calibration methods based on uniaxial tensile testing, in-situ calibration can utilize the built-in deformation measurement function in visuotactile sensors, simplifying the equipment and process. Meanwhile, such processes do not require dismantling the sensor or breaking the elastomer, making it suitable for routine in-situ mechanical calibration of encapsulated sensor products.

### C. Motivation

The principle of existing in-situ calibration methods is to measure the relationship between the normal contact force and the indentation depth (maximum normal displacement) and fit the mechanical parameters of the elastomer. The visuotactile

This work was supported by the National Natural Science Foundation of China under Grant 52375017. *(Corresponding author: Yao Jiang.)*
The authors are with the Institute of Manufacturing Engineering, Department of Mechanical Engineering, Tsinghua University, Beijing 100084, China (e-mail: mingxuan-li@foxmail.com; zlw21@mails.tsinghua.edu.cn; zhouyanh23@mails.tsinghua.edu.cn; litm@mail.tsinghua.edu.cn; jiangyao@mail.tsinghua.edu.cn).

sensor can replace the displacement feedback system, while acquiring accurate contact force still relies on high-performance force/torque sensors. However, commercial force sensors are expensive and fragile. For example, the common-used ATI force/torque sensor can cost tens of thousands of dollars. They rely on fragile strain gauges that must be wrapped in bulky packages [13]. Also, additional accessories, such as a data acquisition board, are needed. Therefore, such methods are usually more suitable for use in laboratories while quite limited in utility and cost-effectiveness in the wild and factories.

In addition, in-situ calibration based on force sensors usually employs a robotic arm to control the pressing action. On the one hand, the displacement precision of the robotic arm is low compared to that of a micro-motion platform. On the other hand, visuotactile sensors have been widely used in areas other than robotic grasping and manipulation, such as microgeometry measurement [14], robotic foot sensing [15], and traction force measurement [16]. Such sensors are used independently without a robotic arm, so no self-contained displacement adjustment function is available.

Due to these issues, we actively consider low-cost and easy-to-use calibration to promote the application of visuotactile sensors in the robotics community. Such calibration methods should avoid the use of expensive force/torque sensors, and have precise displacement adjustment capabilities to avoid the need for a robotic arm. Meanwhile, since visuotactile sensors are used in different scenarios with different forms, the solution that relies on only a single person's easy operation has a high practical value. The realized calibration approach should be as simple as operating a micrometer in the wild with minimal expertise.

### D. Contribution

This article proposes a simple and low-cost method and the related device, EasyCalib, for rough in-situ calibration of visuotactile sensors [see Fig.1]. Experiments showed that EasyCalib could effectively measure the mechanical parameters and benefit the optimization of contact distributed force reconstruction. The main contributions are as follows:

1) We propose a new contact model for in-situ mechanical calibration. Based on the theoretical derivation of contact mechanics and simulation validation, the described method can fit the mechanical parameters by utilizing shape-variable relationships without the need of measuring force/torque-related information.
2) We simplify the in-situ calibration process of visuotactile sensors. During the calibration process, the user only needs to record readings from the displacement platform and the tactile sensor, eliminating the need for other electronics like force/torque sensors.
3) The cost of in-situ calibration devices is reduced. While commercial sensors used for calibration cost thousands of dollars, EasyCalib's parts cost less than $120 and do not require other accessories. The elastic indenters used are simple to manufacture and can be replaced, thus reducing costs and increasing universality.

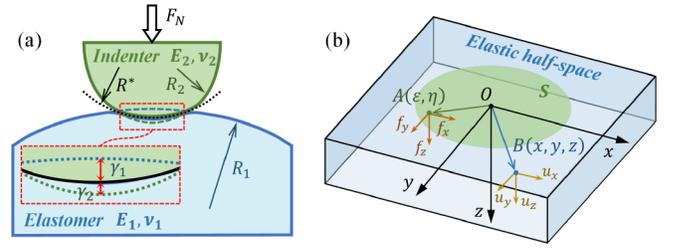

**Fig. 2.** Normal contact model. (b) Torsion contact model.

## II. THEORY

### A. Deformation-based calibration

For two elastomers in normal contact, the external loads on them satisfy the same distribution. Let the deformation fields of them be $D_1$ and $D_2$, respectively. Once the structure and material of them are determined, the mapping relations can be constructed with an appropriate force reconstruction mode as

$$F = g_1(E_1, \nu_1, D_1) = g_2(E_2, \nu_2, D_2). \quad (1)$$

We denote that the first is the sensor's elastomer while the second is a standard elastic calibration indenter ($E_2$ and $\nu_2$ are known). Therefore, the mechanical parameters $E_1$ and $\nu_1$ of the elastomer can be solved by measuring $D_1$ and $D_2$. In this article, two contact scenarios are considered to construct the calibration maneuver similar to the traditional indentation test.

### B. Normal Contact Model

As shown in Fig. 2(a), we consider a soft elastomer with mechanical parameters of $E_1$, $\nu_1$, and an effective radius of $R_1$ in pure normal contact with a hemispherical elastic indenter with parameters of $E_2$, $\nu_2$, and $R_1$. According to Hertz's contact theory [18], by using the assumption of elastic half-space, frictionless, and small contact radius, the maximum displacement on the surface can be calculated as

$$\gamma_i = \frac{1 - \nu_i^2}{E_i} \cdot \frac{E^*}{R^*} a^2, \quad i \in \{1, 2\}, \quad (2)$$

$$\text{where} \quad \frac{1}{R^*} = \frac{1}{R_1} + \frac{1}{R_2}, \quad \frac{1}{E^*} = \frac{1}{E_1^*} + \frac{1}{E_2^*}, \quad \frac{1}{E_i^*} = \frac{1 - \nu_i^2}{E_i}, \quad (3)$$

and $a$ denotes the contact radius. According to Finan et al. [17], the positive error due to the parabolic approximation can be canceled out by the negative error processed by the large strain effect. Therefore, the violation of the small strain assumption still fulfills the requirements. Studies have demonstrated that Hertz's theory can be effectively extended to cases even if the relevant assumptions do not strictly hold [19], [20]. Regardless of the indenter's geometry, Poisson's ratio, or interfacial friction, as long as the soft elastomer is thick enough, the error of the load is less than 4.4%, even at indentations up to 20% [21].

However, Hertz's theory is still sensitive to the violation of the elastic half-space assumption and is highly sensitive [17], [21]. To solve this problem, we introduce Tatara's theories [22], [23] to correct the results. The modified displacement field is expressed as:

$$u_i = (u_i)_H - (u_i)_W, \quad i \in \{1, 2\}, \tag{4}$$

where $(u_i)_H$ denotes the result of Hertz's theory, and $(u_i)_W$ denotes the displacement of the contact surface expansion caused by the displacement of the medium in the range $z > 2R_i$, which should be deducted from the total displacement. According to the explicit expression of Tatara's theory [24]:

$$\gamma_i = \frac{3}{4E_i^*} \cdot \frac{F}{a} - \frac{(1+\nu_i)(3-2\nu_i)}{4\pi E_i R_i} F, \tag{5}$$

while Eq. (2) still stands. According to Eq. (2) and Eq. (5),

$$\sqrt[2]{C(B_2\gamma_1 - B_1\gamma_2)} = \sqrt[3]{C(A_2\gamma_1 - A_1\gamma_2)}, \tag{6}$$

where $A_i = \dfrac{E^*}{E_i^* R^*}$, $B_i = \dfrac{(1+\nu_i)(3-2\nu_i)}{3\pi E_i R_i} \dfrac{E^*}{R^*}$,

$$C = (A_1 B_2 - A_2 B_1)^{-1}. \tag{7}$$

Since the elastomer of a visuotactile sensor can be modeled as a cylinder with the equivalent thickness [8], it should be satisfied that $R_1 \gg R_2$. Besides, the indenter is artificially set to be stiffer than the soft elastomer, i.e., $E_2^* > E_1^*$. Thus,

$$A_1 > A_2, \quad B_1 \ll B_2, \quad C > 0, \tag{8}$$

and Eq. (6) can be expressed as

$$C(B_2\gamma_1 - B_1\gamma_2)^3 = (A_2\gamma_1 - A_1\gamma_2)^2. \tag{9}$$

Considering that Tatara's correction term is a small quantity compared to Hertz's contact term, and the size of the elastomer is larger than that of the indenter, we have

$$A_1 > A_2 \gg B_2 \gg B_1. \tag{10}$$

Therefore, the higher order terms of $B_1$ in Eq. (9) can be neglected, and Eq. (9) can be rewritten as

$$\gamma_2 = K_2 \cdot \gamma_1^2 + K_1 \cdot \gamma_1 + \varphi(\gamma_1), \tag{11}$$

where $\varphi^2 + (K_2 + K_3)\gamma_1^2 \varphi + K_2 \gamma_1^3 (K_3 \gamma_1 - K_1) = 0, \tag{12}$

and $K_1 = \dfrac{A_2}{A_1}$, $K_2 = \dfrac{B_2^2(A_1 B_2 - 3A_2 B_1)}{A_1^2 A_2 (A_1 B_2 - A_2 B_1)}$,

$$K_3 = \frac{B_2^3}{A_1 A_2 (A_1 B_2 - A_2 B_1)}. \tag{13}$$

When the contact has just occurred, $a = 0$ (i.e., $\gamma_1 = \gamma_2 = 0$). Expanding Eq. (9) with implicit function derivation yields

$$\left.\frac{d\gamma_2}{d\gamma_1}\right|_{a=0} = \frac{A_2}{A_1} = \frac{1-\nu_2^2}{1-\nu_1^2} \cdot \frac{E_1}{E_2}, \tag{14}$$

$$\gamma_2|_{a \to 0} = K_2 \cdot \gamma_1^2|_{a \to 0} + K_1 \cdot \gamma_1|_{a \to 0}. \tag{15}$$

According to Eq. (15), $\varphi$ is positive. Also, according to Eq. (10), $K_1 \gg K_3 > K_2$ stands. Therefore, solving Eqs. (11) and (12) yields ($\Lambda$ denotes the high-order minors)

$$\gamma_2 = K_1 \cdot \gamma_1 + \sqrt{K_1 K_2} \cdot \gamma_1^{1.5}$$
$$-K_4 \cdot \gamma_1^2 + \frac{K_4^2}{2\sqrt{K_1 K_2}} \cdot \gamma_1^{2.5} + \Lambda, \tag{16}$$

where $K_4 = \dfrac{K_3 - K_2}{2} = \dfrac{3B_1 B_2^2}{2A_1^2(A_1 B_2 - A_2 B_1)}. \tag{17}$

Consider a suitable set of parameters: $E_1 = 0.1\text{MPa}$, $E_2 = 0.2\text{MPa}$, $\nu_1 = \nu_2 = 0.4$, $R_1 = 50\text{cm}$, $R_2 = 5\text{cm}$. Then we have (dimensionless): $A_1 = 0.1232$, $A_2 = 0.0616$, $B_1 = 7.5319 \times 10^{-4}$, $B_2 = 3.7760 \times 10^{-3}$, and it supports the approximate relationship in Eq. (10). In this case,

$$\gamma_2 = 0.5 \cdot \gamma_1 + 0.0847 \cdot \gamma_1^{1.5} - 0.0025 \cdot \gamma_1^2$$
$$+ 3.776 \times 10^{-5} \cdot \gamma_1^{2.5} + \Lambda. \tag{18}$$

The last two items can be omitted, considering computational efficiency. Therefore, Eq. (16) can be approximated as

$$\gamma_2 = K_1 \cdot \gamma_1 + \sqrt{K_1 K_2} \cdot \gamma_1^{1.5} + \Lambda. \tag{19}$$

The above derivation assumes that the contact forces satisfy the Hertz's pressure distribution, and calculate the displacement form based on this assumption. However, the transverse displacement due to extrusion has not been considered. Yoffe pointed out that the transverse displacement leading to surface displacement would be slightly deeper indentation than that given by the Hertz's theory [25]. Since the size of indenter is much closer to the radius of the contact region than that of the elastomer, the influence of transverse displacement on $A_2$ is significant and cannot be ignored. According to Yoffe's theory, $A_2$ should be corrected as

$$A_2' = \frac{32}{9\pi} \frac{E^*}{E_2^* R^*} = \frac{32}{9\pi} A_2. \tag{20}$$

Therefore,

$$\gamma_2 = \frac{32}{9\pi} K_1 \cdot \gamma_1 + \sqrt{K_1 K_2'} \cdot \gamma_1^{1.5} + \Lambda, \tag{21}$$

In summary, in order to obtain the first relationship of $E_1$ and $\nu_1$, we can fit the maximum normal displacement values of the elastomer and the indenter using

$$\gamma_2 = H_1 \cdot \gamma_1 + H_2 \cdot \gamma_1^{1.5}, \quad H_1 = \frac{32}{9\pi} \cdot \frac{1-\nu_2^2}{1-\nu_1^2} \cdot \frac{E_1}{E_2}. \tag{22}$$

Considering that $K_2$ is related to the shape of the contact and has a large signal-to-noise ratio in practical applications, $H_2$ is not regarded as useful information.

### C. Torsion Contact Model

As shown in Fig. 2(b), we consider the generalized 2-d contact model in elastic half-space. Take $A(\varepsilon, \eta)$ to be a point on the contact region $S$, and $B(x, y, z)$ to be a point in the elastomer. According to the Boussinesq-Cerritti integral equation [26], the action of the distributed forces $f_x(\varepsilon, \eta)$, $f_y(\varepsilon, \eta)$, and $f_z(\varepsilon, \eta)$ can be described by a potential function that satisfies the Laplace equation:

$$F_m = -\iint_S f_m(\varepsilon, \eta)[z \ln(\rho + z) - \rho], \quad m \in \{x, y, z\}, \tag{23}$$

where $\rho = \sqrt{(\varepsilon - x)^2 + (\eta - y)^2 + z^2}. \tag{24}$

According to Love's method [26], the displacements at any point on the contact surface ($z = 0$) can be expressed as

$$\begin{cases} u_x = \dfrac{1}{4\pi G}\left[\left(2\dfrac{\partial^2 F_x}{\partial z^2} - \dfrac{\partial^2 F_z}{\partial x \partial z}\right) + 2\nu\dfrac{\partial \psi}{\partial x}\right] \\ u_y = \dfrac{1}{4\pi G}\left[\left(2\dfrac{\partial^2 F_y}{\partial z^2} - \dfrac{\partial^2 F_z}{\partial y \partial z}\right) + 2\nu\dfrac{\partial \psi}{\partial y}\right] \end{cases}, \quad (25)$$

$$\text{where } \psi = \dfrac{\partial F_x}{\partial x} + \dfrac{\partial F_y}{\partial y} + \dfrac{\partial F_z}{\partial z}, \quad (26)$$

and $G = E/2(1 + \nu)$ represents the shear modulus.

Under normal contact, we consider a hemispherical elastic indenter relative to an elastomer that then undergoes a torsion with the angle of $\theta_{\text{total}}$, perpendicular to the normal direction. The curl of the elastomer's deformation field satisfies

$$\text{rot}(\boldsymbol{u}_1) = \dfrac{\partial u_{1,y}}{\partial x} - \dfrac{\partial u_{1,x}}{\partial y}$$
$$= \dfrac{1}{4\pi G_1}\left(\dfrac{\partial H_y}{\partial x} - \dfrac{\partial H_x}{\partial y}\right) + \dfrac{\nu_1}{2\pi G_1}\left(\dfrac{\partial^2 \psi}{\partial x \partial y} - \dfrac{\partial^2 \psi}{\partial y \partial x}\right), \quad (27)$$

$$\text{where } H_x = 2\dfrac{\partial^2 F_x}{\partial z^2} - \dfrac{\partial^2 F_z}{\partial x \partial z}, \quad H_y = 2\dfrac{\partial^2 F_y}{\partial z^2} - \dfrac{\partial^2 F_z}{\partial y \partial z}. \quad (28)$$

The forces and the potential functions of two objects at the same contact position should be equal in value and opposite in sign. Thus, the curl of the elastomer's deformation satisfies

$$\text{rot}(\boldsymbol{u}_2) = \dfrac{\partial u_{2,y}}{\partial x} - \dfrac{\partial u_{2,x}}{\partial y}$$
$$= -\dfrac{1}{4\pi G_2}\left(\dfrac{\partial H_y}{\partial x} - \dfrac{\partial H_x}{\partial y}\right) + \dfrac{\nu_2}{2\pi G_2}\left(\dfrac{\partial^2 \psi}{\partial x \partial y} - \dfrac{\partial^2 \psi}{\partial y \partial x}\right), \quad (29)$$

According to our previous work [27], the relative rotation angle $\theta_{\text{total}}$ of the sticking region with respect to the relative rotation angle $\theta_{\text{stick}}$ of the elastomer can be expressed as

$$\theta_{\text{total}} = -\left(\dfrac{G_1}{G_2} + 1\right)\theta_{\text{stick}} + \dfrac{L_{xy} - L_{yx}}{2}. \quad (30)$$

Previously, we did not give specific expressions for $L_{xy}$ and $L_{yx}$. By combining Eqs. (27) and (29), we can obtain

$$L_{xy} - L_{yx} = \int \dfrac{\nu_1 - \nu_2}{8\pi G_2}\left(\dfrac{\partial^2 \psi}{\partial x \partial y} - \dfrac{\partial^2 \psi}{\partial y \partial x}\right) dt, \quad (31)$$

The partial derivatives of a potential function should satisfy continuity. Thus, $L_{xy} - L_{yx} = 0$ stands, and

$$\left|\dfrac{\theta_{\text{total}}}{\theta_{\text{stick}}}\right| = \dfrac{\theta_2 + \theta_1}{\theta_1} = \dfrac{G_1}{G_2} + 1, \quad (32)$$

where $\theta_1$ and $\theta_2$ denote the rotation angles of the contact surface with respect to the elastomer and the indenter.

The potential theory relies on strict elastic half-space assumptions. Due to the small size of the elastic indenter and violation of the large thickness assumption, its rotation angle should be corrected. According to the Jäger theory [28], the rotation angle $\theta_2$ satisfies the relationship with the torque $M$ concerning the neutral axis (small angle approximation):

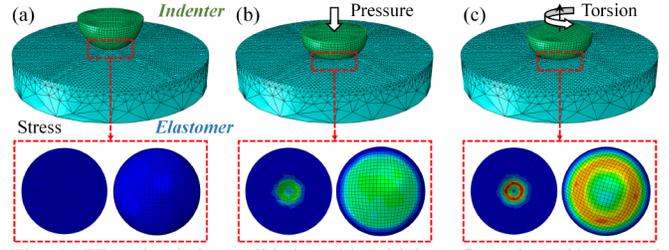

**Fig. 3.** FEA simulation validation. (a) Initial configuration. (b) Normal pressure. (c) Torsion.

$$M = \dfrac{16 G_i a^3}{3}\theta_i, \quad i \in \{1, 2\}, \quad (33)$$

where $a$ denotes the contact radius. It is generalized to the case we discussed due to the fulfillment of Eq. (32). However, a sphere's deformation occurs near the contact region and involves deeper layers. This issue cannot be neglected for elastic indenters with small dimensions. Therefore, we attempt to generalize the idea of rod model [29] to the torsion contact case. Let the indentation depth be $\Delta h$, and the indenter can be equated to a cylinder of thickness $h^* = R - \Delta h$. Based on the material mechanics analysis, the additional angle $\Delta \theta_2$ satisfies

$$M = \dfrac{\pi G_2 R_2^4}{2(R_2 - \Delta h)}\Delta\theta_2. \quad (34)$$

Due to the adjoint transverse displacement, the equivalent contact radius can be approximated as

$$a = \left(1 + \dfrac{\Delta h}{R_2}\right) \cdot \sqrt{R_2^2 - (R_2 - \Delta h)^2}, \quad (35)$$

Substituting Eq. (35) into Eqs (33) and (34) yields

$$\dfrac{\Delta\theta_2^*}{\theta_2} = \dfrac{32}{3\pi}(1-\beta)(2\beta-\beta^2)^{1.5}(1+\beta)^3, \quad \beta = \dfrac{\Delta h}{R_2}. \quad (36)$$

During the calibration, we artificially select that $\beta = 0.2$ (i.e., the indentation depth is 0.2 times the radius). Therefore,

$$\dfrac{\theta_2^*}{\theta_2} = \dfrac{\theta_2 + \Delta\theta_2^*}{\theta_2} \approx 2.014, \text{ and } \dfrac{\theta_2^*}{\theta_1} = 2.014 \cdot \dfrac{G_1}{G_2} \quad (37)$$

In summary, in order to obtain the second relationship of $E_1$ and $\nu_1$, we can fit the maximum torsion angle values of the elastomer and the indenter using the following relationship

$$\theta_2 = H_3 \cdot \theta_1, \quad H_3 = 2.014 \cdot \dfrac{1 + \gamma_2}{1 + \gamma_1} \cdot \dfrac{E_1}{E_2}, \quad (38)$$

Thus, $E_1$ and $\nu_1$ can be calculated from Eqs. (22) and (38):

$$E_1 = \left[0.993 - 0.279 \cdot \dfrac{H_3}{H_1}(1 - \nu_2)\right] \cdot \dfrac{H_3 E_2}{1 + \nu_2}, \quad (39)$$

$$\nu_1 = 1 - 0.562 \cdot \dfrac{H_3}{H_1}(1 - \nu_2). \quad (40)$$

III. SIMULATION VALIDATION

We used the standard FEA simulations to verify the described contact theories. As shown in Fig. 3(a), the soft elastomer and elastic indenter were modeled in the Abaqus engine. The elastomer was modeled as a cylinder with a radius of 15 mm and a thickness of 5 mm, and the elastic indenter as a hemisphere with a radius of 5 mm. Based on the actual parameters of the Tac3D sensor [30], the Young's modulus of the elastomer was selected as $E_1 = 1$MPa and Poisson's ratio as $v_1 = 0.48$. During the interaction, the friction coefficient of the two parts was set to be 0.4, with the penalty friction model, and the linear-elastic model, geometrical nonlinearities, and finite slip were considered. The bottom of the elastomer and the indenter were constrained to simulate that both were fixed on two mutually parallel platforms. In addition, the mesh close to the surface is refined to reduce errors and avoid divergence.

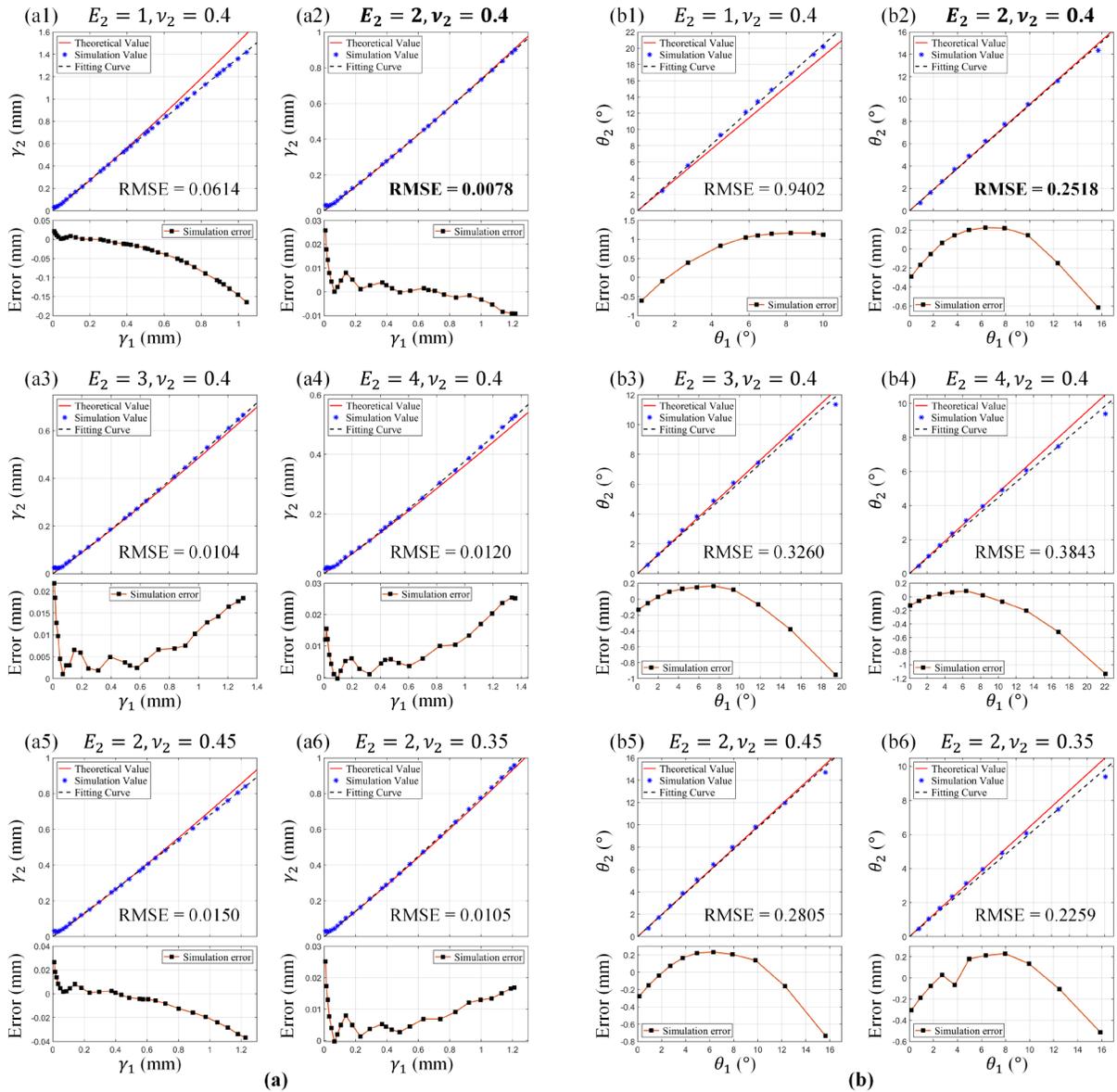

**Fig. 4.** Simulation results. (a) Validation of normal contact model. (b) Validation of torsion contact model. For each subgraph, the upper half is the comparison between theoretical and simulated values, and the lower half is the error curve.

TABLE I
CALIBRATION ERRORS OF DIFFERENT TEST CASES

| Relative error | $E_2 = 1$MPa, $v_2 = 0.4$ | $E_2 = 2$MPa, $v_2 = 0.4$ | $E_2 = 3$MPa, $v_2 = 0.4$ | $E_2 = 4$MPa, $v_2 = 0.4$ | $E_2 = 2$MPa, $v_2 = 0.45$ | $E_2 = 2$MPa, $v_2 = 0.35$ |
|---|---|---|---|---|---|---|
| $H_1$ | 14.13% | 5.05% | 1.24% | **0.13%** | 8.27% | 1.86% |
| $H_3$ | 7.22% | 0.89% | 3.46% | 6.28% | 1.16% | **0.75%** |
| $E_1$ | 9.50% | **1.07%** | 1.89% | 4.26% | 1.87% | 3.22% |
| $v_1$ | 6.55% | 6.11% | **5.03%** | 6.65% | 9.42% | 8.56% |

The contact process was divided into two stages. In the first stage, the normal force $F_N = 20N$ was set to be loaded downward from the top center of the elastic indenter, as shown in Fig. 3(b). The accuracy of the normal contact theory was verified by comparing the maximum normal displacements ($\gamma_1$ and $\gamma_2$). In the second stage, the indentation depth was adjusted to 1 mm ($\beta = 0.2$), and the torque $M = 15N \cdot mm$ perpendicular to the normal direction was set at the same position [see Fig. 3(c)], and the accuracy of the torsion contact theory was verified by comparing the maximum rotation angles ($\theta_1$ and $\theta_2$). In this case, the rotation angles were obtained by fitting the point cloud attached to the model at the contact position. Since the loading in Abaqus consists of multiple steps, we can export the field output in different steps to analyze the contact deformation of the whole process.

As shown in Fig. 5, $E_2$ and $\nu_2$ of the indenter are varied respectively to compare the theoretical and simulation results of normal and torsional contact. Results indicate that the proposed theory can effectively estimate displacement and rotation. The $\gamma_1$-$\gamma_2$ curves obtained by Eq. (22) and the $\theta_1$-$\theta_2$ curves obtained by Eq. (38) show good agreement with the simulation results, and the best results are obtained in the case of $E_2 = 2MPa$ and $\nu_2 = 0.4$. For normal contact, the errors decrease and then increase with the increase of contact displacement (or rotation): when the deformation is small, the system noise leads to large relative errors; when the shape variable is large, the influence of the nonlinear effect increases and exceeds the applicability of the proposed theory. The results above indicate that the mid-range displacement value (0.2-0.8 mm) should be selected in real calibration. In addition, the variations of $\gamma_1$-$\gamma_2$ curves and $\theta_1$-$\theta_2$ curves exhibit the same characteristics with the changes of $E_2$ and $\nu_2$:

1) As $E_2$ increases, the errors first decrease and then increase, and the optimal result is taken when $E_2 = 2E_1 = 2MPa$ [see Figs. 4(a1)-4(a4) and 4(b1)-4(b4)]. When the value of $E_2$ is close to $E_1$, the premise of the modeling that the indenter is stiffer than the soft elastomer is violated, resulting in low predicted displacements. Also, when the value of $E_2$ is much larger than $E_1$, the difference between the contact characteristics of the two parts increases, resulting in high influence of the higher-order nonlinear terms.

2) The variation of $\nu_2$ has a relatively small impact on the theoretical error [see Figs. 4(a1), 4(a5)-4(a6), and 4(b1), 4(b5)-4(b6)]. As $\nu_2$ increases (or decreases), the theoretical curve still exhibits the same downward (or upward) shift as the simulation curve with a small error. However, considering the silicone's material properties, the range of values for $\nu_2$ is usually small. Therefore, compared to $E_2$, v2 has a smaller contribution to fitting, and the main factor affecting calibration performance is the selection of the indenter's Young's modulus.

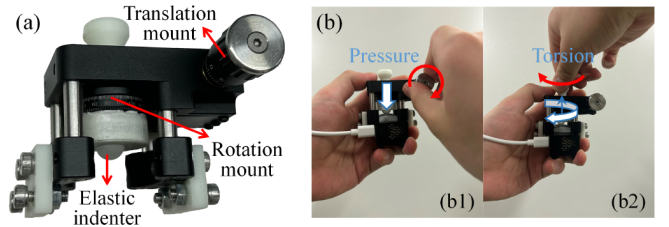

Fig. 5. (a) Design and structure of EasyCalib. (b) Usage of EasyCalib.

TABLE II
COMPONENT AND APPROXIMATE COST LIST

| Component | Cost |
| --- | --- |
| Translation mount (resolution of 0.001mm) | $50 |
| Rotation mount (resolution of 2°) | $37 |
| 4 construction rods (6mm) | $9 |
| 2 mounting brackets | $9 |
| 3D printed pieces (photopolymer) | $2 |
| Elastic indenter (PDMS) | $2 |
| Screws (M4) | $1 |
| Total | $110 |

Table I demonstrates the deviation of the parameters obtained by fitting from the theoretical values. $H_1$ depends on both $E_2$ and $\nu_2$, and the relative error of $H_1$ is smaller when $E_2$ is larger; and $H_3$ is mainly affected by $E_2$, and it is more accurate the closer $E_2$ is to $2E_1$. Under both influences, the calibration of $E_1$ and $\nu_1$ achieves the best results in the second and third test cases, respectively. Therefore, the proposed theory has good validity and accuracy when the selected mechanical parameters are $E_2 = 2E_1 \sim 3E_1$ and $\nu_2$ is near 0.4.

## IV. DEVICE AND EXPERIMENT

### A. Design and usage of EasyCalib

Based on the above theories and verifications, we propose EasyCalib, a deformation-based calibration device, and the related method. As shown in Fig. 5 (a), EasyCalib consisted of four main components: elastic indenter, translation mount, rotation mount, and mounting bracket [see Table II]. Except for translation and rotation mounts that could be purchased from instrument manufacturers and standard components such as screws, all other components could be 3D printed on consumer-grade equipment and completed within half an hour. Compared to commercial F/T sensors, EasyCalib was several orders of magnitude cheaper (the part cost is only $110), providing an affordable and user-friendly calibration solution.

A characteristic of this scheme lies in balancing operational simplicity and usage standardization. In application, the tactile sensor can be fixed to EasyCalib through the mounting bracket, while the contact between the indenter and the elastomer can be controlled through translation and rotation mounts [see Fig. 5(b)]: the micrometer that controls the translation mount can adjust the normal contact by rotating, while torsional contact can be achieved by rotating the handle that controls the rotation mount. The calibration process is as follows:

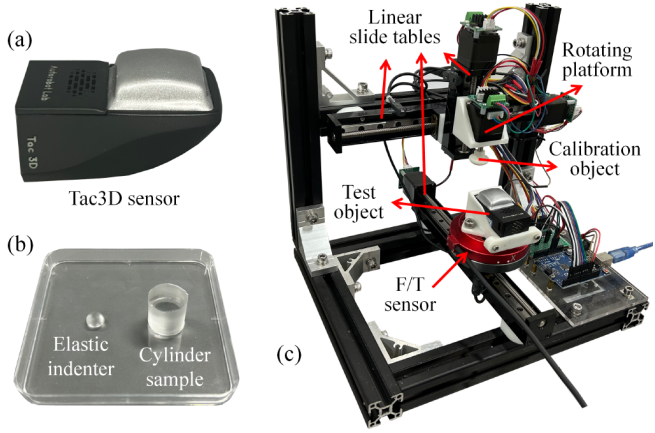

**Fig. 6.** (a) Tac3D sensor. (b) Elastic indenter and cylinder sample. (c) Experiment setup for normal contact and torsion contact calibration.

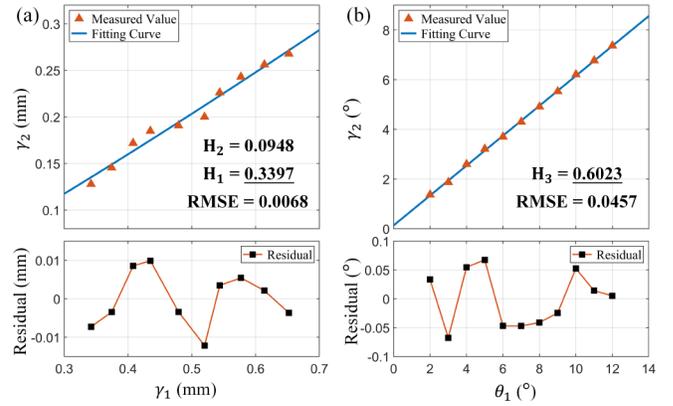

**Fig. 7.** In-situ calibration evaluation. (a) Results of normal contact. (b) Results of torsion contact.

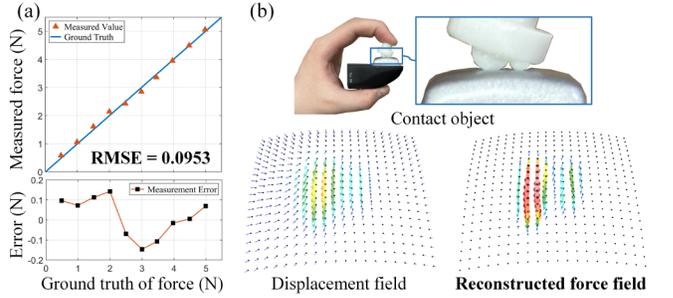

**Fig. 8.** Force reconstruction evaluation (within 0~5N). (a) Quantitative evaluation results. (b) Qualitative evaluation of reconstructed force field.

1) Adjust the micrometer, and record a series of total displacement $\gamma_1 + \gamma_2$ and the maximum indentation $\gamma_1$ measured by the sensor, and fit $H_1$ according to Eq. (22).
2) Set $\gamma_2 = 1mm$, adjust the handle, and record a series of total rotational angles $\theta_1 + \theta_2$ and the rotational angle $\theta_1$ measured by the sensor. Then, fit $H_3$ according to Eq. (38). Here, we suggest using our previous algorithm [27] to implement rotation measurement in the tactile sensor.
3) $E_1$ and $\nu_1$ can be calculated according to Eqs. (39), (40).

### B. In-situ Calibration Evaluation

We chose the Tac3D sensor [30] as the calibration model. The productized Tac3D, as shown in Fig. 6(a), has accurate and reliable visuotactile measurements, and its internal structure is the same as the previously proposed OneTip [31]. Based on experience, the Young's modulus of the Tac3D product is about 0.22~0.38 MPa, and the Poisson's ratio is about 0.38~0.42. Therefore, we try to change the stiffness of the elastic indenter to comply with the parameter values suggested in Section III. As shown in Fig. (b), the elastic indenter and the cylindrical sample (as a standard part) were made from the same batch of poly-di-methyl-siloxane (PDMS) using the same process. According to Fuard et al. [32], the mechanical parameters of PDMS can be adjusted by varying the cross-linker concentration and baking condition. We explored different preparation processes and measured the mechanical parameters of cylindrical samples based on the platform shown in Fig. 6(c). The contact between the calibrator and the test object (cylindrical sample) can be controlled using the slide tables and rotating platform. By reading the force measured by a F/T sensor (with a resolution of 0.01N) with the displacement provided by the slide tables, the standard compression experiments [33] can be executed. Finally, we selected the cross-linker ratio of 1:10 and the baking condition of $70°C$ and 12 hours to prepare the indenter, which gave $E_2 = 1.1035MPa$ and $\nu_2 = 0.3883$.

The calibration procedure was also applied on the platform shown in Fig. 6(c) by replacing the test object with a Tac3D sensor. We used the pipeline introduced in Section IV to perform the normal and torsion contacts. To reduce the random error, we repeated the sampling ten times and averaged the results. The results are shown in Fig. 7. Based on the obtained $\gamma_1$-$\gamma_2$ and $\theta_1$-$\theta_2$ curves, the fitted values of $H_1$ and $H_3$ are 0.3397 and 0.6023, respectively. Among these, the displacement curves' errors are larger than the angular curves. The main reason is that the contribution of the nonlinear term fact caused by the non-zero equivalent radius of Tac3D. Finally, the in-situ calibration of Tac3D according to Eqs. (39) and (40) resulted in $E_1 = 0.3305MPa$ and $\nu_1 = 0.3905$.

### C. Force Reconstruction Evaluation

Due to the difficulty of comparing with real mechanical parameters (with the same reason introduced in [8]), we evaluated the accuracy of EasyCalib by comparing the reconstructed obtained distributed forces. Following our previous approach [11], we imported the measured $E_1$ and $\nu_1$ into the elastomer model built in Abaqus, and obtained the stiffness matrices of the 3D displacements to the 3D distributed forces through finite element simulation. This process typically takes only ten minutes or so. Also, the centralized force can be obtained by integrating the distributed force. Using the platform above, we compared the ground truth measured by the F/T sensor with the normal force reconstructed by Tac3D based on this stiffness matrix, as shown in Fig. 8(a). The quantitative evaluation shows that the force estimation is very close to the ground truth, that is, the root mean square error of the combined force is 0.0953 N, and the maximum error is not more than 0.15 N.

Since measurement errors in evaluating distributed force fields are often unjustified [34], we employed the typical contact case to qualitatively illustrate the accuracy of force distribution estimation, as shown in Fig. (b). The calibrated Tac3D efficiently distinguished multiple contact regions that were not connected and obtained distribution properties that were compatible with the deformation field. The above evaluations demonstrate the accuracy of obtained mechanical parameters and its contribution to force reconstruction.

## V. Conclusion

This article describes an in-situ calibration device, EasyCalib, for the routine measurement of mechanical parameters (Young's modulus and Poisson's ratio) of visuotactile sensors. Based on the derived theories of normal and torsion contact, EasyCalib relies on contact deformation measurements to obtain the relational equations to fit the mechanical parameters, eliminating the need for costly F/T sensors and data processing. With its low-cost design (about $110) and simple calibration process, users can implement it in the wild as easily as operating an angle scale. Simulations and experiments demonstrated the validity and accuracy of the method, and showed its application in optimizing distributed force reconstruction. However, the limitation of the current scheme is the reliance on theories with approximate assumptions, which limits the operational flexibility and accuracy. The future research will actively explore the contribution of data from finite element simulation, priori information, and end-to-end training to the in-situ calibration process, and provide more powerful alternatives as well as new usage cases.

## References

[1] R. S. Dahiya, G. Metta, M. Valle, and G. Sandini, "Tactile sensing: From humans to humanoids," *IEEE Trans. Robot.*, vol. 26, no. 1, pp. 1–20, Feb. 2010.

[2] Q. Li, O. Kroemer, and Z. Su, "A review of tactile information: Perception and action through touch," *IEEE Trans. Robot.*, vol. 36, no. 6, pp. 1–16, Jul. 2020.

[3] S. Zhang et al., "Hardware technology of vision-based tactile sensor: A review," *IEEE Sensors J.*, vol. 22, no. 22, pp. 21410–21427, 2022, doi: 10.1109/JSEN.2022.3210210.

[4] M. Li, L. Zhang, T. Li, and Y. Jiang, "Marker displacement method used in vision-based tactile sensors—from 2D to 3D: A review," *IEEE Sensors J.*, vol. 23, no. 8, pp. 8042–8059, Apr. 2023.

[5] R. Sui, L. Zhang, Q. Huang, T. Li, and Y. Jiang, "A novel incipient slip degree evaluation method and its application in adaptive control of grasping force," *IEEE Trans. Autom. Sci. Eng.*, doi: 10.1109/TASE.2023.3241325.

[6] M. Oller, M. P. i Lisbona, D. Berenson, and N. Fazeli, "Manipulation via membranes: High-resolution and highly deformable tactile sensing and control," in *Proc. Conf. Robot Learn. (CoRL)*, Mar. 2023, pp. 1850–1859.

[7] T. Bi, C. Sferrazza, and R. D'Andrea, "Zero-shot sim-to-real transfer of tactile control policies for aggressive swing-up manipulation," *IEEE Robot. Automat. Lett.*, vol. 6, no. 3, pp. 5761–5768, Jul. 2021.

[8] C. Zhao, J. Ren, H. Yu, and D. Ma, "In-situ mechanical calibration for vision-based tactile sensors," in *Proc. IEEE Int. Conf. Robot. Autom. (ICRA)*, May 2023, pp. 10387–10393.

[9] D. Ma, E. Donlon, S. Dong, and A. Rodriguez, "Dense tactile force estimation using GelSlim and inverse FEM," in *Proc. Int. Conf. Robot. Autom. (ICRA)*, May 2019, pp. 5418–5424.

[10] C. Sferrazza and R. D'Andrea, "Design, motivation and evaluation of a full-resolution optical tactile sensor," *Sensors*, vol. 19, no. 4, p. 928, 2019.

[11] L. Zhang, T. Li, and Y. Jiang, "Improving the force reconstruction performance of vision-based tactile sensors by optimizing the elastic body," *IEEE Robot. Autom. Lett.*, vol. 8, no. 2, pp. 1109–1116, Feb. 2023.

[12] S. Ito, N. Hirai, and Y. Ohki, "Changes in mechanical and dielectric properties of silicone rubber induced by severe aging," *IEEE Trans. Dielectr. Electr. Insul.*, vol. 27, no. 3, pp. 722–730, 2020.

[13] R. Ouyang and R. Howe, "Low-cost fiducial-based 6-Axis force-torque sensor," in *Proc. IEEE Int. Conf. Robot. Autom.*, May 2020, pp. 1653–1659.

[14] M. Johnson, F. Cole, A. Raj, and E. Adelson, "Microgeometry capture using an elastomeric sensor," *ACM Trans. Graph.*, vol. 30, p. 46, 2011.

[15] G. Zhang, Y. Du, Y. Zhang, and M. Y. Wang, "A tactile sensing foot for single robot leg stabilization," in *Proc. IEEE Int. Conf. Robot. Autom. (ICRA)*, May 2021, pp. 14076–14082.

[16] Y. Li et al., "Imaging dynamic three-dimensional traction stresses," *Sci. Adv.*, vol. 8, no. 11, Mar. 2022, Art. no. eabm0984.

[17] C.-E. Wu, K.-H. Lin, and J.-Y. Juang, "Hertzian load–displacement relation holds for spherical indentation on soft elastic solids undergoing large deformations," *Tribol. Int.*, vol. 97, pp. 71–76, May 2016.

[18] A. C. Fischer-Cripps, *Introduction to Contact Mechanics*. New York: Springer-Verlag, 2007, pp. 103–106.

[19] E. Ciulli, A. Betti, and P. Forte, "The applicability of the Hertzian formulas to point contacts of spheres and spherical caps," *Lubricants*, vol 10, no. 10, p. 233, 2022.

[20] Z. Cui et al., "Haptically quantifying Young's modulus of soft materials using a self-locked stretchable strain sensor," *Adv. Mater.*, vol 34, no. 25, 2022, Art. no. 2104078.

[21] J. D. Finan, P. M. Fox, and B. Morrison, "Non-ideal effects in indentation testing of soft tissues," *Biomech. Model. Mechan.*, vol. 13, no. 3, pp.573–584, 2014.

[22] Y. Tatara, "Extensive theory of force-approach relations of elastic spheres in compression and in impact," *J. Eng. Mater. Technol.*, vol. 111, pp. 163–168, 1989.

[23] Y. Tatara, "On compression of rubber elastic sphere over a large range of displacements—Part 1: Theoretical study," *J. Eng. Mater. Technol.*, vol. 113, no. 3, pp. 285–291, 1991.

[24] Z. Wang and X. Liu, "Adhesion of large-deformation elastic spherical contact," *Tribol. Int.*, vol. 119, pp. 559–566, Nov. 2018.

[25] E. H. Yoffe, "Modified hertz theory for spherical indentation," *Philos. Mag. A*, vol. 50, no. 6, pp. 813–828, 1984.

[26] J. L. Johnson, *Contact Mechanics*. Cambridge, U.K.: Cambridge Univ. Press, 1987, pp. 45–50.

[27] M. Li, Y. H. Zhou, T. Li, and Y. Jiang, "Incipient slip-based rotation measurement via visuotactile sensing during in-hand object pivoting," 2023, arXiv:2309.05366.

[28] J. Jäger, "Axi-symmetric bodies of equal material in contact under torsion or shift," *Arch. Appl. Mech.*, vol. 65, pp. 478–487, 1995.

[29] G. S. Boltachev, N. B. Volkov, and N. M. Zubarev, "Tangential interaction of elastic spherical particles in contact," *Int. J. Solids Struct.*, vol. 49, pp. 2107–2114, 2012.

[30] L. Zhang, Y. Wang, and Y. Jiang, "Tac3D: A novel vision-based tactile sensor for measuring forces distribution and estimating friction coefficient distribution," 2022, arXiv:2202.06211.

[31] M. Li, Y. H. Zhou, L. Zhang, T. Li, and Y. Jiang, "OneTip is enough: A non-rigid input device for single-fingertip human-computer interaction with 6-DOF," Feb. 2024. [Online]. Available: https://doi.org/10.36227/techrxiv.170775314.44150885/v1.

[32] D. Fuard, T. Tzvetkova-Chevolleau, S. Decossas, P. Tracqui, and P. Schiavone, "Optimization of poly-di-methyl-siloxane (PDMS) substrates for studying cellular adhesion and motility," *Microelectron. Eng.*, vol. 85, pp. 1289–1293, 2008.

[33] T. K. Kim, J. K. Kim, and O. C. Jeong, "Measurement of nonlinear mechanical properties of pdms elastomer," *Microelectron. Eng.*, vol. 88, pp. 1982–1985, 2011.

[34] L. V. Duong, "BiTac: A soft vision-based tactile sensor with bidirectional force perception for robots," *IEEE Sensors J.*, vol. 23, no. 9, pp. 9158-9167, 2023.